\pdfoutput=1

\documentclass[11pt]{article}

\usepackage{emnlp2021}

\usepackage{times}
\usepackage{latexsym}
\usepackage{booktabs}
\usepackage[T1]{fontenc}

\usepackage[utf8]{inputenc}

\usepackage{microtype}
\usepackage{graphicx}
\usepackage{amsmath, amsthm, amssymb}
\usepackage{multirow, multicol}
\usepackage{pifont}
\usepackage{subfig}
\usepackage{enumitem}
\usepackage{algorithm}
\usepackage[noend]{algpseudocode}
\usepackage{caption} 

\usepackage{comment}
\newcommand\Tstrut{\rule{0pt}{2.6ex}}       
\newcommand\Bstrut{\rule[-0.9ex]{0pt}{0pt}} 
\newcommand{\TBstrut}{\Tstrut\Bstrut} 

\makeatletter
\newcommand{\pushright}[1]{\ifmeasuring@#1\else\omit\hfill$\displaystyle#1$\fi\ignorespaces}
\newcommand{\pushleft}[1]{\ifmeasuring@#1\else\omit$\displaystyle#1$\hfill\fi\ignorespaces}
\makeatother

\usepackage{todonotes}
\usepackage{ulem}

\title{Adversarial Scrubbing of Demographic Information for Text Classification}

\author{Somnath Basu Roy Chowdhury  \qquad
  Sayan Ghosh   \qquad
  Yiyuan Li   \qquad
  {Junier B. Oliva} \\ 
  \texttt{\{somnath, sayghosh, yiyuanli, joliva\}@cs.unc.edu}\AND
  \textbf{Shashank Srivastava}   \qquad
  \textbf{Snigdha Chaturvedi} \\
  \texttt{\{ssrivastava, snigdha\}@cs.unc.edu} \\
  UNC Chapel Hill \\}

\begin{document}
\maketitle
\begin{abstract}

Contextual representations learned by language models can often encode \textit{undesirable attributes}, like demographic associations of the users,  while being trained for an unrelated target task. We aim to scrub such undesirable attributes and learn fair representations while maintaining performance on the target task.  In this paper, we present an adversarial learning framework ``\textbf{Ad}versarial \textbf{S}crubber'' (\textsc{AdS}), to debias contextual representations. We perform theoretical analysis to show that our framework converges without leaking demographic information under certain conditions. We extend previous evaluation techniques by evaluating debiasing performance using Minimum Description Length (MDL) probing. Experimental evaluations on 8 datasets show that \textsc{AdS}  generates representations with minimal information about demographic attributes while being maximally informative about the target task. 


\end{abstract}

\section{Introduction}

Automated systems are increasingly being used for real-world applications like filtering college applications \cite{basu2019predictive}, determining credit eligibility \cite{ghailan2016improving}, making hiring decisions \cite{chalfin2016productivity}, etc. For such tasks, predictive models are trained on data coming from human decisions, which are often biased against certain demographic groups \cite{mehrabi2019survey, blodgett2020language, shah2019predictive}. 
Biased decisions based on demographic attributes can have lasting economic, social and cultural consequences. 

Natural language text is highly indicative of demographic attributes of the author \cite{koppel2002automatically, burger2011discriminating, nguyen2013old, verhoeven2014clips, weren2014examining, rangel2016overview, verhoeven2016twisty, blodgett2016demographic}. Language models can often encode such demographic associations even without having direct access to them. Prior works have shown that intermediate representations in a deep learning model encode demographic associations of the author or person being spoken about \cite{blodgett2016demographic, elazar2018adversarial, elazar2021amnesic}. Therefore, it is important to ensure that decision functions do not make predictions based on such representations. 


In this work, we focus on removing demographic attributes encoded in data representations during training text classification systems. To this end, we present ``\textbf{Ad}versarial \textbf{S}crubber'' (\textsc{AdS}) to remove information pertaining to \textit{protected attributes} (like gender or race) from intermediate representations during training for a \textit{target task} (like hate speech detection).  
Removal of such features ensures that any prediction model built on top of those representations will be agnostic to demographic information during decision-making.

\textsc{AdS} can be used as a plug-and-play module during training any text classification model 
to learn fair intermediate representations.  The framework consists of 4 modules: 
Encoder, Scrubber, Bias discriminator and Target classifier.  The Encoder generates contextual representation of an input text. Taking these encoded contextual representations as input, the Scrubber tries to produce fair representations for the target task. The Bias discriminator and Target classifier predict the protected attribute and target label respectively  from the Scrubber's output.  
The framework is trained end-to-end in an adversarial manner \cite{goodfellow2014generative}. 

We provide theoretical analysis to show  that under certain conditions Encoder and Scrubber converge without leaking information about the protected attribute.  We evaluate our framework on {5} dialogue datasets, 2 Twitter-based datasets and a Biographies dataset with different target task and protected attribute settings. We extend  previous evaluation methodology for debiasing by measuring Minimum Description Length (MDL) \cite{voita2020information} of labels given representations, instead of probing accuracy. 
MDL provides a finer-grained evaluation benchmark for measuring debiasing performance.  We compute MDL using off-the-shelf classifiers\footnote{We use MLPClassifier modules from \href{https://scikit-learn.org/stable/}{scikit-learn}.} making it easier to reproduce. Upon training using \textsc{AdS} framework, we observe a significant gain in MDL for protected attribute prediction as compared to fine-tuning for the target task. 
Our contributions are:
\begin{itemize}[noitemsep, topsep=0.3pt]
    \setlength\itemsep{0.3em}
    \item We present \textbf{Ad}versarial \textbf{S}crubber (\textsc{AdS}), an adversarial framework to learn fair representations for text classification.
    \item We provide theoretical guarantees to show that Scrubber and Encoder converge without leaking demographic information.
    \item We extend previous evaluation methodology for adversarial debiasing by framing performance in terms of MDL.
    \item Experimental evaluations on 8 datasets show that models trained using \textsc{AdS} generate representations where probing networks achieve near random performance on protected attribute inference while performing similar to the baselines on target task.
    \item We show that \textsc{AdS} is scalable and can be used to remove multiple protected attributes simultaneously.
\end{itemize}


\section{Related Work}
Contextual representations learned during training for a target task can be indicative of features unrelated to the task. 
Such representations can often encode undesirable demographic attributes, as observed in unsupervised word embeddings \cite{bolukbasi} and sentence embeddings \cite{may2019measuring}. Prior work has analysed bias in different NLP  systems like machine translation \cite{park2018reducing, stanovsky2019evaluating, font2019equalizing, saunders-byrne-2020-reducing}, NLI \cite{rudinger2017social}, text classification \cite{dixon2018measuring, kiritchenko2018examining, sap2019risk, liu2021authors}, language generation \cite{sheng2019woman} among others. 

Debiasing sensitive attributes for fair classification was introduced as an optimization problem by \citet{zemel2013learning}. Since then, adversarial training \cite{goodfellow2014generative} frameworks have been explored for protecting sensitive attributes for NLP tasks \cite{zhang2018mitigating, li2018towards, elazar2018adversarial, liu2020mitigating}. 

Our work is most similar to \citet{elazar2018adversarial}, which achieves \textit{fairness by blindness} by learning intermediate representations which are oblivious to a protected attribute. 
We compare the performance of \textsc{AdS} with \citet{elazar2018adversarial} in our experiments.

\begin{figure}[t!]
	\centering
 	\includegraphics[width=0.37\textwidth]{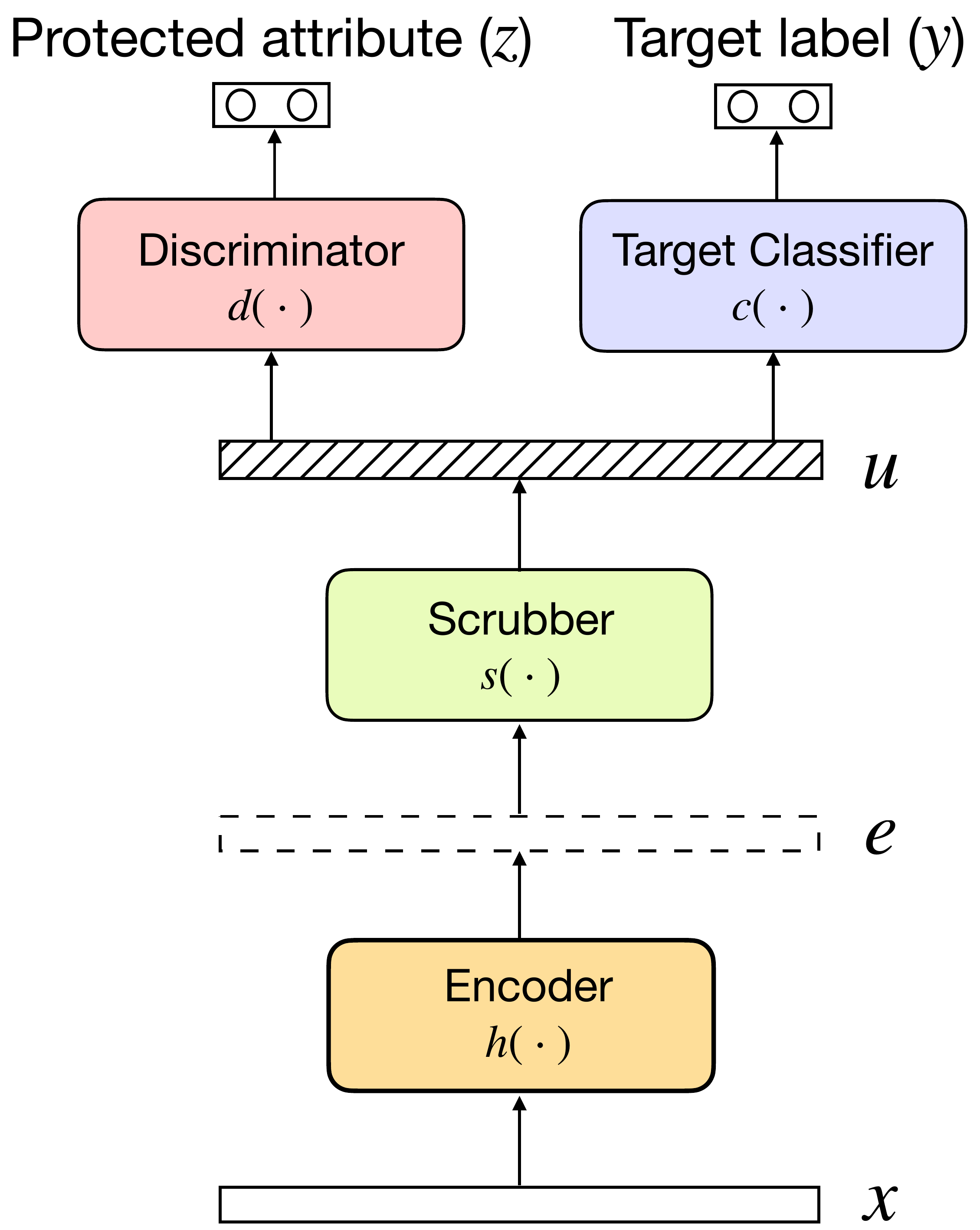}
	\caption{Architecture of the Adversarial Scrubber (\textsc{AdS}). Encoder receives an input $x$ to produce $e$. Scrubber uses $e$ to produce $u$. Bias discriminator $d$ and Target classifier $c$ infer protected attribute $z$ and target task label $y$ from $u$.}
	\label{fig:dynamic-model}
\end{figure}

\begin{algorithm*}[t!]
	\caption{\textsc{AdS} Training algorithm}
	\begin{algorithmic}[1]
	\For{number of training iterations}
		\State Sample a minibatch $\{x_{i}, y_i, z_i\}^{m}_{i=1} \sim \mathcal{D}$ 
		\State  Bias discriminator $d$ is updated using the gradients:
		\begin{equation}
		    \nabla_{\theta_d} \frac{1}{m}\sum\limits_{i=1}^{m} \mathcal{L}_d (d(u_i), z_i)
		    \label{eqn:disc-grad}
		\end{equation}
		\State Update the Encoder $h$, Scrubber $s$, and Task Classifier $c$ using the gradients:
		\begin{equation}
		\begin{aligned}
		\nabla_{\theta_c, \theta_s, \theta_h} \frac{1}{m}\sum\limits_{i=1}^{m} \bigg[\mathcal{L}_c(c(u_i), y_i) -\lambda_1 H(d(u_i)) + \lambda_2 \delta(d(u_i))\bigg]
		\end{aligned}
		\label{eqn:gen-grad}
		\end{equation}
	\EndFor
\end{algorithmic}
	\label{alg:training}
\end{algorithm*} 

\section{Adversarial Scrubber}
 
\textsc{AdS} takes text documents $\{x_1, x_2, \ldots, x_n\}$ as input from a dataset $\mathcal{D}$  with corresponding target labels $\{y_1, y_2, \ldots, y_n\}$. Every input $x_i$ is also associated with a \textit{protected attribute} $z_i \in \{1, 2, ... K\}$. 
Our goal is to construct a model $f(x)$ such that it doesn't rely on $z_i$ while making the prediction $y_i = f(x_i)$.
The framework consists of 4 modules
: (i) Encoder $h(\cdot)$ with weights $\theta_h$, (ii) Scrubber $s(\cdot)$ with weights $\theta_s$, (iii) Bias discriminator $d(\cdot)$ with weights $\theta_d$ and (iv) Target classifier $c(\cdot)$ with weights $\theta_c$ as shown in Figure~\ref{fig:dynamic-model}. The Encoder receives a text input $x_i$, and produces an embedding $e_i = h(x_i)$,  which  is forwarded to the Scrubber. The goal of the Scrubber is to produce representation $u_i = s(h(x_i))$, such that $y_i$ can be easily inferred from $u_i$ by the Target classifier, $c$, but $u_i$ does not have the information required to predict the protected attribute $z_i$ by the Bias discriminator $d$. Our setup also includes a Probing network $q$, which helps in evaluating the fairness of the learned representations.

In the rest of this section, we describe \textsc{AdS} assuming a single Bias discriminator. However, \textsc{AdS} can easily be extended to incorporate multiple discriminators for removing several protected attributes (discussed in Section~\ref{sec:multi-disc}).

\noindent\textbf{Scrubber}: The Scrubber receives  the  input 
representation $h(x_i)$  from Encoder and generates representation $u_i = s(h(x_i))$. The goal of the Scrubber is to produce representations such that the Bias discriminator finds it difficult to predict the protected attribute $z_i$. To this end, we consider two loss functions:\\[-1em]

\noindent\underline{Entropy loss}: In the Entropy loss, the Encoder and Scrubber parameters are jointly optimized to \textit{increase} the entropy of the prediction probability distribution, $H(d(u_i))$.\\[-1em] 

\noindent\underline{$\delta$-loss}: 
The $\delta$-loss function penalizes the model if the discriminator assigns a high probability to the correct protected-attribute class. For every input instance, we form an output mask $m_i \in \mathbb{R}^{1 \times K}$ where $K$ is the number of \textit{protected attribute classes}. $m_i^{(k)} = 1$ if $z_i = k$ and $0$ otherwise. 
The Encoder and Scrubber \textit{minimizes} 
the $\delta$-loss 
defined as:
\begin{equation}
\delta(d(u_i)) = m_i^T  \mathrm{softmax}_{gumble}(d(u_i))
\label{delta}
\end{equation}

\noindent where $\mathrm{softmax}_{gumble}(\cdot)$ is the gumble softmax function \cite{jang2016categorical}. In our experiments, we use a combination of the entropy and $\delta$ losses.\\[-1em]

\noindent\textbf{Target classifier}: The Target classifier predicts the target label $y_i$ from $u_i$ by optimizing the cross entropy loss: $\mathcal{L}_c(c(u_i), y_i)$. 

The Scrubber, Target classifier, and Encoder parameters are updated simultaneously to minimize the following loss:
\begin{equation}
	\begin{aligned}
	\mathcal{L}_s(e_i, y_i) = \mathcal{L}_c(c(u_i), y_i) -\lambda_1 H(d(u_i)) \\+ \lambda_2 \delta(d(u_i))
	\end{aligned}
    \label{eqn:gen-loss}
\end{equation}
\noindent where $\lambda_1$ and $\lambda_2$ are positive hyperparameters.\\[-1em]

\noindent\textbf{Bias discriminator}: The Bias discriminator, which predicts the protected attribute $z_i$, is trained to reduce the cross-entropy loss for predicting $z_i$ denoted as $\mathcal{L}_d (d(u_i), z_i)$. The discriminator output  is $d(u_i) \in \mathbb{R}^{K}$, where $K$ is the number of protected attribute classes.\\[-1em]

\noindent\textbf{Training}: The Bias discriminator and Scrubber (along with Target classifier and Encoder) are trained in an iterative manner as shown in Algorithm~\ref{alg:training}. 
First, the Bias discriminator is updated  using gradients from the loss in Equation~\ref{eqn:disc-grad}.  Then, the Encoder, Scrubber and Target classifier are updated simultaneously using the gradients shown in Equation~\ref{eqn:gen-grad}. \\[-1em]

\noindent\textbf{Probing Network}: 
\citet{elazar2018adversarial} showed that in an adversarial setup even when the discriminator achieves random performance for predicting $z$, it is still possible to retrieve $z$ using a separately trained classifier. 
Therefore, to evaluate the amount of information related to $y$ and $z$ present in representations $u$, we use a probing network $q$. 
{After \textsc{AdS} is trained, we train $q$ on representations $h(x)$ and $s(h(x))$, to predict $y$ and $z$ ($q$ is trained to predict $y$ and $z$ separately). We consider an \textit{information leak} from a representation, if $z$ can be predicted from it with above random  performance. If the prediction performance of $q$ for $z$ is significantly above the random baseline, it means that there is information leakage of the protected attribute and it is not successfully  guarded. }


\section{Theoretical Analysis} 

\noindent\textbf{Proposition 1.} \textit{Minimizing $\mathcal{L}_s$ is equivalent to increasing Bias discriminator loss $\mathcal{L}_d$.} \\[-0.5em]


\noindent \underline{Proof}: Entropy and $\delta$-loss components of $\mathcal{L}_s$ tries to increase the bias discriminator loss. The discriminator cross-entropy loss $\mathcal{L}_d$  can be written as:

\begin{equation}
    \begin{aligned}
        \mathcal{L}_d(v_i, o_i) &= H(v_i, o_i) \\
        &= D_{KL}(v_i, o_i) + H(v_i)
    \end{aligned}
    \label{eqn:prop-1-1}
\end{equation}

\noindent where $o_i = d(u_i)$, the Bias discriminator output probability distribution and $v_i$ is a one-hot target distribution $\{v_i \in \mathbb{R}^{K}, v_i^k = 1 \lvert z_i = k\}$. 
As $H(o_i)$ increases (Equation~\ref{eqn:gen-loss}), $D_{KL}(o_i, v_i)$ value also increases (since $v_i$ is a one-hot vector), thereby increasing $\mathcal{L}_d(o_i, v_i)$ (in Equation~\ref{eqn:prop-1-1}). Therefore, $\mathcal{L}_d$ increases as we minimize the Scrubber loss component $-H(o_i)$. 

The same holds true for the $\delta$-loss component. $\delta(o_i)$ reduces the probability assigned to the true output class which increases the cross entropy loss $\mathcal{L}_d$ (detailed proof provided in Appendix~\ref{sec:prop1} due to space constraint). Minimizing the entropy and $\delta$-loss components of the Scrubber loss $\mathcal{L}_s$ increases $\mathcal{L}_d$ for a fixed Bias discriminator. Therefore, assuming our framework converges to $(\theta_s^*, \theta_h^*, \theta_d^*)$ using gradient updates    from $\mathcal{L}_s$ we have:

\begin{equation}
    \mathcal{L}_d(\theta_s^*, \theta_h^*, \theta_d^*) \geq \mathcal{L}_d(\theta_s, \theta_h, \theta_d^*)
    \label{eqn:prop-1}
\end{equation}

\noindent where $(\theta_s, \theta_h)$ can be any Scrubber and Encoder parameter setting.

\vspace*{5pt} 
\noindent\textbf{Proposition 2.}  \textit{Let the discriminator loss $\mathcal{L}_d$ be convex in $\theta_d$, and continuous differentiable for all $\theta_d$. Let us assume the following:}

\noindent\textit{ (a)  $\theta_h^{(0)}$ and $\theta_s^{(0)}$ are Encoder and Scrubber parameters when the Scrubber output representation $s(h(x))$ does not have any information about $z$ (one trivial case would be when $s(h(x)) = \vec{0}$, if $\theta_s = \vec{0} \lor \theta_h = \vec{0}$)}.

\noindent\textit{ (b) $\theta_d^{(0)}$ minimizes $\mathcal{L}_d$ when $s(h(x))$ does not have any information about $z$ (this is achieved when $d(\cdot)$ always predicts the majority baseline for $z$). $\forall (\theta_s, \theta_h)$, the following holds true:} 
\begin{equation*}
    \small
    \mathcal{L}_d(\theta_s, \theta_h, \theta_d^{(0)}) = \mathcal{L}_d(\theta_s^{(0)}, \theta_h^{(0)}, \theta_d^{(0)})
\end{equation*}

\noindent\textit{(c) the adversarial framework converges with parameters $\theta_s^*, \theta_h^*$ and $\theta_d^*$.}

\noindent \textit{Then, $\mathcal{L}_d(\theta_s^*, \theta_h^*, \theta_d^{*}) = \mathcal{L}_d(\theta_s^*, \theta_h^*, \theta_d^{(0)})$ which implies that the Bias discriminator loss does not benefit from updates of $\theta_s$ and $\theta_h$.}\\[-0.5em]

\noindent \underline{Proof}: As the Bias discriminator converges to $\theta_d^*$, we have:
\begin{equation}
    \mathcal{L}_d(\theta_s^*, \theta_h^*, \theta_d^*)   \leq \mathcal{L}_d(\theta_s^*, \theta_h^*, \theta_d^{(0)})
    \label{eqn:disc-conv}
\end{equation}

\noindent$\theta_h$ and $\theta_s$ are updated using gradients from $\mathcal{L}_s$ (Equation~\ref{eqn:gen-loss}). Since the Encoder and the Scrubber parameters converge to $\theta_h^*$ and $\theta_s^*$ respectively, from {Proposition 1} (Equation~\ref{eqn:prop-1}) we have:
\begin{equation}
    \mathcal{L}_d(\theta_s^*, \theta_h^*, \theta_d^*)   \geq \mathcal{L}_d(\theta_s^{(0)}, \theta_h^{(0)}, \theta_d^{*})
    \label{eqn:gen-conv}
\end{equation}

\noindent We can show that:
\begin{equation}
	\begin{aligned}
		\mathcal{L}_d(\theta_s^*, \theta_h^*, &\theta_d^{(0)}) 
		\\&\geq \mathcal{L}_d(\theta_s^*, \theta_h^*, \theta_d^*) \hspace{1.4cm}\hfill{\text{ (Equation~\ref{eqn:disc-conv})}}\\
		&\geq \mathcal{L}_d(\theta_s^{(0)}, \theta_h^{(0)}, \theta_d^*) \hspace{1cm}\hfill{\text{ (Equation~\ref{eqn:gen-conv})}}\\
		&\geq \mathcal{L}_d(\theta_s^{(0)}, \theta_h^{(0)}, \theta_d^{(0)}) \hspace{0.3cm}\hfill\text{ (Assumption 2b)}\\
		&= \mathcal{L}_d(\theta_s^{*}, \theta_h^{*}, \theta_d^{(0)}) \hspace{0.8cm}\hfill\text{ (Assumption 2b)}\\
	\end{aligned}
\end{equation}

\noindent Therefore, $\mathcal{L}_d(\theta_s^*, \theta_h^*, \theta_d^*)  =\mathcal{L}_d(\theta_s^{*}, \theta_h^{*}, \theta_d^{(0)})$.

\vspace*{5pt} 
\noindent \textbf{Proposition 3.} \textit{Let us assume that the Bias discriminator $d(\cdot)$ is strong enough to achieve optimal accuracy of predicting $z$ from $s(h(x))$ and assumptions in Proposition 2 hold true. Then, Encoder and Scrubber converge to ($\theta_h^*, \theta_s^*$) without leaking information about the protected attribute $z$.}\\[-0.5em]

\noindent \underline{Proof}: An optimal Bias discriminator $d(\cdot)$ minimizes the prediction entropy, thereby increasing the entropy and $\delta$-loss. Given $(\theta_h^{(0)}, \theta_s^{(0)})$, the Scrubber loss $\mathcal{L}_s$ is maximized for an optimal $\theta_d^{(0)}$ (From Proposition 1, $\mathcal{L}_s(\theta_s^{(0)}, \theta_h^{(0)}, \theta_d^{(0)})   \geq \mathcal{L}_s(\theta_s^{(0)}, \theta_h^{(0)}, \theta_d)$, since $\mathcal{L}_d$ is decreasing with $\delta(o_i)$ and $-H(o_i)$). Then, for any other discriminator $\theta_d^*$ we have:
\begin{equation}
\mathcal{L}_s(\theta_s^{(0)}, \theta_h^{(0)}, \theta_d^{*}) \leq \mathcal{L}_s(\theta_s^{(0)}, \theta_h^{(0)}, \theta_d^{(0)}) 
\label{eqn:gen-loss-d}
\end{equation}
 
\noindent Following assumption 2b,do where $\theta_d^{(0)}$ is the optimal Bias discriminator we can show that: 
\begin{equation}
	\begin{aligned}
		\mathcal{L}_s(\theta_s^{(0)}, \theta_h^{(0)}, &\theta_d^{(0)}) 
	\\&\geq \mathcal{L}_s(\theta_s^{(0)}, \theta_h^{(0)}, \theta_d^{*}) \hspace{0.5cm}\hfill{\text{(Equation~\ref{eqn:gen-loss-d}})}
		\\& \geq \mathcal{L}_s(\theta_s^{*}, \theta_h^{*}, \theta_d^{*}) \hspace{0.7cm}\hfill{\text{($\mathcal{L}_s$ converges)}}
	\end{aligned}
	\label{eqn:converge}
\end{equation}

\noindent Therefore, $\mathcal{L}_s(\theta_s^{*}, \theta_h^{*}, \theta_d^{*}) \leq \mathcal{L}_s(\theta_s^{(0)}, \theta_h^{(0)}, \theta_d^{(0)})$.

\noindent From $(\theta_s^{(0)}, \theta_h^{(0)}, \theta_d^{(0)})$, our framework converges to $(\theta_s^{*}, \theta_h^{*}, \theta_d^{*})$ as the Scrubber loss $\mathcal{L}_s$ decreases (Equation~\ref{eqn:converge}). Then, from Proposition 2 we have
$$\mathcal{L}_d(\theta_s^*, \theta_h^*, \theta_d^*)  \geq \mathcal{L}_d(\theta_s^{(0)}, \theta_h^{(0)}, \theta_d^{(0)}) $$

\noindent As $\mathcal{L}_d$ does not decrease, and $d(\cdot)$ is optimal it shows that no additional information about $z$ is revealed which the Bias discriminator can leverage to reduce $\mathcal{L}_d$. This shows that starting from $(\theta_s^{(0)}, \theta_h^{(0)}, \theta_d^{(0)})$ where assumptions in Proposition 2 hold, our framework converges to $(\theta_s^{*}, \theta_h^{*}, \theta_d^{*})$ without  revealing information about $z$. 

\section{Experiments}

In this section, we describe our experimental setup and evaluate \textsc{AdS} on several benchmark datasets.

\begin{table}[t!]
	\centering
	\resizebox{0.32\textwidth}{!}{
		\begin{tabular}{l |c c c}
	\toprule[1pt]
	\multirow{2}{*}{\textsc{Dataset}} &  \multicolumn{3}{c}{Split} \TBstrut\\ 
	& Train & Dev & Test\TBstrut\\
	\hline
	Funpedia & 24K & 2.9K & 2.9K \Tstrut\\
	Wizard & 3.5K & 0.1K & 0.1K \\
	ConvAI2 & 69K & 4.5K & 4.5K \\
	LIGHT & 38K & 2.2K & 4.5K \\
	OpenSub  & 210K & 25K & 29K \\
	\hline
	DIAL &  166K & - & 151K\Tstrut\\
	PAN16 &  160K & - & 9K\\
	PAN16 &  160K & - & 10K\Bstrut\\
	\hline
	Biographies & 257K & 40K & 99K\TBstrut\\
	\hline
\end{tabular}
	}
	\caption{ Dataset statistics. }
	\label{tab:data_stats}
\end{table}

\subsection{Dataset} 
We evaluate \textsc{AdS} on 5 dialogue datasets, 2 Twitter-based datasets and a Biographies dataset. 


\noindent(a) \textbf{Multi-dimensional bias in dialogue systems}: 
We evaluate \textsc{AdS} on 5 dialogue datasets: Funpedia, ConvAI2, Wizard, LIGHT and OpenSub,  introduced by \citeauthor{dinan2020multi} (\citeyear{dinan2020multi}). 
These datasets are annotated with multi-dimensional gender labels: the gender of the person being {spoken about}, the gender of the person being {spoken to}, and gender of the {speaker}. We consider the \textit{gender} of the person being \textit{spoken about} as our {protected attribute}. The {target task} in our setup is \textit{sentiment classification}. For obtaining the target label, we label all instances using the rule-based  sentiment classifier VADER \cite{hutto2014vader}, into three classes: positive, negative and neutral. The dialogue datasets: Funpedia, Wizard, ConvAI2, LIGHT and OpenSub were downloaded from ``md\_gender'' dataset in huggingface library.\footnote{https://huggingface.co/datasets/md\_gender\_bias} We use the same data split provided in huggingface for these dataset.

\noindent (b) \textbf{Tweet classification}: We experiment on two Twitter datasets. 
First, we consider the DIAL dataset \cite{blodgett2016demographic}, where each tweet is annotated with ``\textit{race}'' information of the author, which is our {protected attribute} and the {target task} is \textit{sentiment classification}. We consider two race categories: non-Hispanic blacks and whites. 
Second, we consider the PAN16 \cite{rangel2016overview} dataset where each tweet is annotated with the author's \textit{age} and \textit{gender} information both of which are {protected attributes}. The {target task} is \textit{mention detection}. We use the implementation\footnote{https://github.com/yanaiela/demog-text-removal} of \citet{elazar2018adversarial} to annotate both datasets.

\begin{figure*}[t!]
	\centering
	\includegraphics[width=\textwidth]{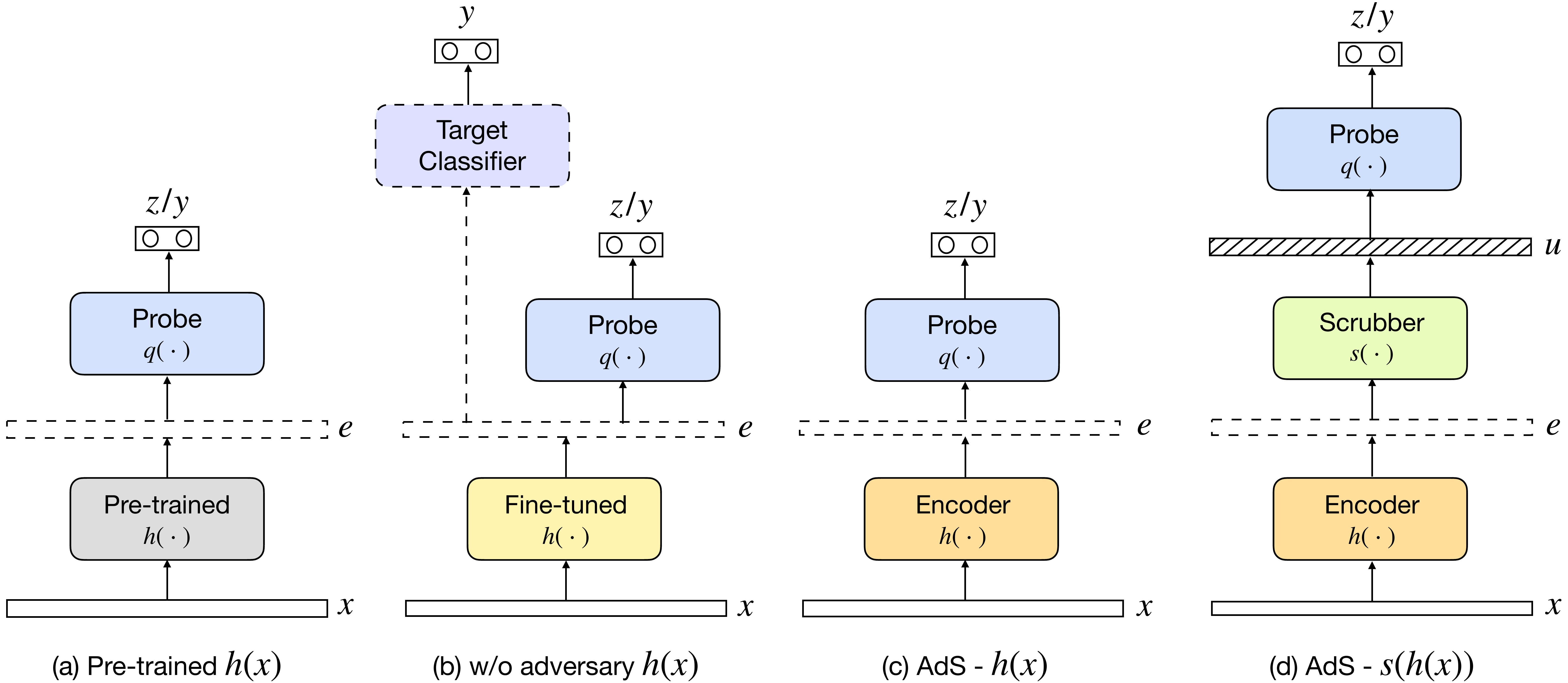}
		\caption{Evaluation setup. We evaluate the performance of the probing network on 4 different  representations. (a) {Pre-trained $h(x)$} obtained using pre-trained Encoder (b) w/o adversary $h(x)$ when the Encoder $h$ was fine-tuned  on the target task (c) \textsc{AdS} $h(x)$ Encoder embeddings and (d) \textsc{AdS} - $s(h(x))$ embeddings from the Scrubber are representations obtained from \textsc{AdS}.}
	\label{fig:eval}
\end{figure*}

\noindent (c) \textbf{Biography classification}: 
We evaluate \textsc{AdS} on biographies dataset \cite{de2019bias}. The target task involves \textit{classification of biographies} into 28 different profession categories, and protected attribute is the \textit{gender} of the person. The dataset has been downloaded and processed from this open-sourced project.\footnote{https://github.com/Microsoft/biosbias} We use the same train-dev-test split of 65:10:25 as the authors. 

 All datasets used in our experiments are balanced. The dataset statistics are reported in Table~\ref{tab:data_stats}.

\begin{table}[t!]
	\centering
	\resizebox{0.5\textwidth}{!}{
		\begin{tabular}{l c c c c c}
	\toprule[1pt]
	\textsc{Dataset}  & $z$ & $y$ & Epoch & $\lambda_1$ & $\lambda_2$\TBstrut\\ 
	\midrule[1pt]
	Funpedia  & Gender (3) & Sentiment (3) & 2 & 1 & 1\Tstrut\\
	Wizard & Gender (2) & Sentiment (3) & 3 & 1 & 0\\
	ConvAI2 & Gender (2) & Sentiment (3) & 1 & 1 & 0\\
	LIGHT  & Gender (2) & Sentiment (3) & 2 & 1 & 0\\
	OpenSub & Gender (2) & Sentiment (3) & 2 & 1 & 0\Bstrut\\
    \hline
    DIAL & Race (2) & Sentiment (2) & 8 & 10 & 0\Tstrut\\
    PAN16 & Gender (2) & Mention (2) & 5 & 10 & 0\\
    PAN16 & Age (2) & Mention (2) & 3 & 10 & 0 \Bstrut\\
    \hline
    Biographies & Gender (2) & Occupation (28) & 2 & 10 & 0\TBstrut\\
	\bottomrule[1pt]
\end{tabular}
	}
	\caption{Hyperparameter settings. Each entry for $z$/$y$ are shown the format ``Attribute Name ($c$)'', where $c$ is the number of classes for that attribute.}
	\label{tab:setup}
\end{table}

\subsection{Implementation details} 
 We use a 2-layer feed-forward neural network with ReLU non-linearity as our Scrubber network $s$. We use BERT-base \cite{devlin2018bert} as our Encoder $h$. Bias discriminator $d$ and Target classifier $c$ take the pooled output of BERT \texttt{[CLS]} representation followed by a single-layer neural network.  All the models were using AdamW optimizer with a learning rate of $2 \times 10^{-5}$. Hyperparameter details for different datasets are mentioned in Table~\ref{tab:setup}. $z$ and $y$ sections in the table report the protected attribute and the target task for each dataset. For each task we also report the number of output classes in paranthesis (e.g. Sentiment (3)). The implementation of this project is publicly available here: \href{https://github.com/brcsomnath/adversarial-scrubber}{https://github.com/brcsomnath/AdS}.

\subsection{Evaluation Framework}
\label{sec:eval-setup}
{In our experiments, we compare representations obtained from $4$ different settings as shown in Figure~\ref{fig:eval}. Figure~\ref{fig:eval}(a), (b) and (c) are our baselines.}
In Figure~\ref{fig:eval}(a), we retrieve $h(x)$ from pre-trained BERT model. In Figure~\ref{fig:eval}(b), we retrieve $h(x)$ from BERT fine-tuned on the target task. In Figure~\ref{fig:eval}(c), Encoder output $h(x)$ from \textsc{AdS} is evaluated. {In Figure \ref{fig:eval}(d), Scrubber output, $s(h(x))$ is evaluated. This represents our final setup \textsc{AdS} - $s(h(x))$.} 

\subsection{Metrics}

{We report the F1-score (F1) of the probing network for each evaluation. However, previous work has shown that probing accuracy is not a reliable metric to evaluate the degree of information related to an attribute encoded in representations \cite{hewitt2019designing}. Therefore, we also report Minimum Description Length (MDL) \cite{voita2020information} of labels given representations. MDL captures the amount of effort required by a \textit{probing network} to achieve a certain accuracy. Therefore, it provides a finer-grained evaluation benchmark which can even differentiate between probing models with comparable accuracies. We compute the online code \cite{rissanen1984universal} for MDL. In the online setting, blocks of labels are encoded by a probabilistic model iteratively trained on incremental blocks of data (further details about MDL is provided in Appendix A.1). We compute MDL using sklearn's  MLPClassifier
\footnote{We use default hyperparameters from \href{https://scikit-learn.org/stable/}{scikit-learn}} at timesteps corresponding to 0.1\%, 0.2\%, 0.4\%, 0.8\%, 1.6\%, 3.2\%, 6.25\%, 12.5\%, 25\%, 50\% and 100\% of each dataset as suggested by \citet{voita2020information}. A higher MDL signifies that more effort is required to achieve the probing performance. Hence,  we expect the debiased representations to have higher MDL for predicting $z$ and a lower MDL for predicting $y$.}



\section{Results}
The evaluation results for all datasets are reported in Table~\ref{tab:dynamic-results}.  For all datasets, we report performances in 4 settings described in Section~\ref{sec:eval-setup}.

\begin{table*}[t!]
	\centering
	\resizebox{\textwidth}{!}{
		\begin{tabular}{ l | c c | c c | c c | c c | c c| c c}
	\toprule[1pt]
	  \multicolumn{1}{c|}{Dataset $\rightarrow$} & \multicolumn{4}{c|}{\textsc{Funpedia}} & \multicolumn{4}{c|}{\textsc{Wizard}} & \multicolumn{4}{c}{\textsc{ConvAI2}}\Tstrut\\ 
	 \multicolumn{1}{c|}{Setup $\downarrow$} & \multicolumn{2}{c}{Gender ($z$)} & \multicolumn{2}{c|}{Sentiment ($y$)} & \multicolumn{2}{c}{Gender ($z$)} & \multicolumn{2}{c|}{Sentiment ($y$)} & \multicolumn{2}{c}{Gender ($z$)} & \multicolumn{2}{c}{Sentiment ($y$)} \\ 
	& \multicolumn{1}{c}{F1 $\downarrow$} & \multicolumn{1}{c}{MDL $\uparrow$} & \multicolumn{1}{c}{F1 $\uparrow$} & \multicolumn{1}{c|}{MDL$\downarrow$} & \multicolumn{1}{c}{F1 $\downarrow$} & \multicolumn{1}{c}{MDL $\uparrow$} & \multicolumn{1}{c}{F1 $\uparrow$} & \multicolumn{1}{c|}{MDL $\downarrow$} & \multicolumn{1}{c}{F1 $\downarrow$} & \multicolumn{1}{c}{MDL $\uparrow$} & \multicolumn{1}{c}{F1 $\uparrow$} & \multicolumn{1}{c}{MDL $\downarrow$}  \Bstrut\\
	\midrule[1pt]
	Random 						& 33.3 & - & 33.3 & - & 50.0 & - & 33.3 & - & 50.0 & - & 33.3 & -  \TBstrut\\
	 Pre-trained $h(x)$  & 56.8 &  24.7 & 62.3 & 46.3 & 78.6 & 3.8 & 46.5 & 7.6 & 80.3 & 100.6 & 62.7 & 133.7\\
	w/o adversary $h(x)$  & 51.0 & 30.9 & \textbf{92.8} & \textbf{2.8} & 67.4 & 5.2 & 85.1  &  {0.2} & 72.8 & 109.0 & {95.6} & \textbf{6.5}\\
	\textsc{AdS} - $h(x)$  & 44.1 & 35.4 & {90.3} & {10.3} & 63.4 & 6.5 & 88.1 & 0.3 & 58.3 & \textbf{134.0} & 95.3 & {10.9}\\
    \textsc{AdS} - $s(h(x))$ & \textbf{29.8}  & \textbf{41.4} & 90.2 & {10.8}& \textbf{54.7} & \textbf{6.9} & \textbf{93.2} & {0.2} & \textbf{56.0} & 133.5 & {95.3}  & 11.0\Bstrut\\
	\bottomrule[1pt]
\end{tabular}

	}
    \resizebox{\textwidth}{!}{
			\begin{tabular}{ l | c c | c c | c c | c c | c c | c c}
		\toprule[1pt]
		\multicolumn{1}{c|}{Dataset $\rightarrow$} & \multicolumn{4}{c|}{\textsc{Light}} & \multicolumn{4}{c|}{\textsc{OpenSub}} & \multicolumn{4}{c}{\textsc{Biographies}}\Tstrut\\ 
		\multicolumn{1}{c|}{Setup $\downarrow$} & \multicolumn{2}{c}{Gender ($z$)} & \multicolumn{2}{c|}{Sentiment ($y$)} & \multicolumn{2}{c}{Gender ($z$)} & \multicolumn{2}{c|}{Sentiment ($y$)} & \multicolumn{2}{c}{Gender ($z$)} & \multicolumn{2}{c}{Occupation ($y$)} \\ 
		& \multicolumn{1}{c}{F1 $\downarrow$} & \multicolumn{1}{c}{MDL $\uparrow$} & \multicolumn{1}{c}{F1 $\uparrow$} & \multicolumn{1}{c|}{MDL $\downarrow$} & \multicolumn{1}{c}{F1 $\downarrow$} & \multicolumn{1}{c}{MDL $\uparrow$} & \multicolumn{1}{c}{F1 $\uparrow$} & \multicolumn{1}{c|}{MDL $\downarrow$} & \multicolumn{1}{c}{F1 $\downarrow$} & \multicolumn{1}{c}{MDL $\uparrow$} & \multicolumn{1}{c}{F1 $\uparrow$} & \multicolumn{1}{c}{MDL $\downarrow$} \Bstrut\\
		\midrule[1pt]
		Random 						& 50.0 & - & 33.3 & - & 50.0 & - & 33.3 & - & 50.0 & - & 3.6 & - \TBstrut\\
		Pre-trained $h(x)$  & 78.6 & 47.1 & 60.5 & 88.7 & 72.3 & 192.4 & 63.9 & 426.2 & 99.2 & 27.6 & 74.3 & 499.9\\
		w/o adversary $h(x)$  & 75.3 & 55.9 & 91.4 & \textbf{8.2} & 70.2 & 311.9 & \textbf{97.5} & \textbf{25.1} & 62.3 & 448.9 & {99.9} & \textbf{2.2}\\
		\textsc{AdS} - $h(x)$  & 60.4 & 73.8 & 92.2 & 16.7 & 40.7  & 371.9 & 96.9 & 37.4 & 62.1 & 444.7 & 99.9 & 3.0\\
		\textsc{AdS} - $s(h(x))$ & \textbf{52.8} & \textbf{74.7} & {92.3}  & {16.4} & {40.7} & \textbf{373.7} & {96.9} & {37.1} & \textbf{57.1} & \textbf{449.5} & 99.9 & 3.3\Bstrut\\
		\bottomrule[1pt]
\end{tabular} 
    }
    
	\resizebox{\textwidth}{!}{
		\begin{tabular}{ l | c c | c c | c c | c c | c c | c c}
		\toprule[1pt]
		\multicolumn{1}{c|}{Dataset $\rightarrow$} & \multicolumn{4}{c|}{\textsc{Dial}} & \multicolumn{8}{c}{\textsc{Pan16}} \Tstrut\\ 
		\multicolumn{1}{c|}{Setup $\downarrow$} & \multicolumn{2}{c}{Race ($z$)} & \multicolumn{2}{c|}{Sentiment ($y$)} & \multicolumn{2}{c}{Gender ($z$)} & \multicolumn{2}{c|}{Mention ($y$)} & \multicolumn{2}{c}{Age ($z$)} & \multicolumn{2}{c}{Mention ($y$)}\\ 
		& \multicolumn{1}{c}{F1 $\downarrow$} & \multicolumn{1}{c}{MDL $\uparrow$} & \multicolumn{1}{c}{F1 $\uparrow$} & \multicolumn{1}{c|}{MDL $\downarrow$} & \multicolumn{1}{c}{F1 $\downarrow$} & \multicolumn{1}{c}{MDL $\uparrow$} & \multicolumn{1}{c}{F1 $\uparrow$} & \multicolumn{1}{c|}{MDL $\downarrow$} & \multicolumn{1}{c}{F1 $\downarrow$} & \multicolumn{1}{c}{MDL $\uparrow$} & \multicolumn{1}{c}{F1 $\uparrow$} & \multicolumn{1}{c}{MDL $\downarrow$} \Bstrut\\
		\midrule[1pt]
		Random & 50.0 & - & 50.0 & - & 50.0 & - & 50.0 & - & 50.0 & - & 50.0 & -  \TBstrut\\
		Pre-trained $h(x)$ & 74.3 & 242.6 & 63.9 & 300.7 &  60.9 & 300.5 & 72.3 & 259.7 & 57.7 & {302.0} & 72.8 & 262.6 \\
		w/o adversary $h(x)$  & 81.7 & 176.2 & \textbf{76.9}   & 99.0 & 68.6 & 267.6 & 89.7 & \textbf{4.0} & 59.0 & 295.4 & {89.3} & \textbf{4.8}\\
		\textsc{AdS} - $h(x)$  & 69.7 & 273.0  & 72.4 & 51.0 & 62.3 & 304.2 & 89.7 & {7.1} & 62.4 & 302.8 & 89.3 & {5.3}\\
		\textsc{AdS} - $s(h(x))$ & {\textbf{58.2}} & \textbf{290.6} & 72.9 & \textbf{56.9} & \textbf{48.6} & \textbf{313.9} & {89.7} & 7.6 & \textbf{50.5} & \textbf{315.1} & {89.2} & 6.0 \Bstrut\\
		\bottomrule[1pt]
\end{tabular}
		\label{tab:tab3}
	}
	\caption{Evaluation results for all datasets. Expected trends for a metric are shown in $\uparrow$- higher scores and $\downarrow$- lower scores. Statistically significant best probing performances for $z$ (lowest F1/highest MDL) and $y$ (highest F1/lowest MDL) are in bold.\footnote{MDL is not a normalized measure and varies depending on the size of the dataset.} \textsc{AdS} - $s(h(x))$ performs the best in guarding information leak of $z$ for all datasets. }
	\label{tab:dynamic-results}
\end{table*}

\noindent\textbf{Dialogue and Biographies dataset}: First, we focus on the results on the  dialogue and biographies datasets reported in Table~\ref{tab:dynamic-results} (first two rows). We observe the following: 
(i) for pre-trained $h(x)$, MDL of predicting $z$ is lower than $y$ for these datasets. This means that information regarding $z$ is better encoded in the pre-trained $h(x)$, than the target label $y$. 
(ii) In ``w/o adversary $h(x)$'' setup, the Encoder is fine-tuned on the target task (without debiasing), upon which MDL for $y$ reduces significantly (lowest MDL achieved in this setting for all datasets) accompanied by a rise in MDL for $z$. However, it is still possible to predict $z$ with a F1-score significantly above the random baseline, 
(iii) ``\textsc{AdS} - $h(x)$'' setup achieves similar F1 score for predicting $y$, but still has a F1-score for $z$ significantly above the random baseline.
(iv) ``\textsc{AdS} - $s(h(x))$'' performs the best in terms of guarding the protected attribute $z$ (lowest prediction F1-score and highest MDL) by achieving near random F1-score across all datasets. It is also able to maintain performance on the target task, as we observe only a slight drop compared to the fine-tuning performance (``w/o adversary $h(x)$'' for predicting $y$).


\begin{table}[t!]
	\centering
	\resizebox{0.5\textwidth}{!}{
        \begin{tabular}{l| c c |c c| c c}
	\toprule[1pt]
	 \multirow{3}{*}{\textsc{Setup}} & \multicolumn{2}{c|}{\textsc{Dial}} & \multicolumn{4}{c}{\textsc{Pan16}}\Tstrut\\
	 & \multicolumn{2}{c|}{Race ($z$)} & \multicolumn{2}{c|}{Gender ($z$)} & \multicolumn{2}{c}{Age ($z$)}\\
	 & $\Delta z$ & Acc$_y$ & $\Delta z$ & Acc$_y$ & $\Delta z$ & Acc$_y$ \Bstrut\\
	\midrule[1pt]
	w/o adversary LSTM & 14.5 & 67.4 & 10.1 & 77.5 & 9.4 & 74.7 \Tstrut\\
	\citet{elazar2018adversarial} & \textbf{4.8} & 63.8 & \underline{4.1} & {74.3} & \underline{5.7} & 70.1\Bstrut\\
	\hline
	w/o adversary BERT & 31.2 & \textbf{76.4} & 18.5 & \underline{89.7} & 10.1 & \textbf{89.3} \Tstrut\\
	\textsc{AdS} - $s(h(x))$ & \underline{8.2}  & \underline{72.9} & \textbf{0.8} & \textbf{89.8} & \textbf{4.7} & \underline{89.2}\\
	\bottomrule[1pt]
\end{tabular}
	}
	\vspace{0.2cm}
	\caption{Comparing \textsc{AdS} with existing baseline. The best and second best performances are in bold and underlined respectively. \textsc{AdS} - $s(h(x))$ achieve the best performance on both settings in the PAN16 dataset and is able to reduce $\Delta_z$ better than baseline on DIAL.}
	\label{tab:baseline}
\end{table}

\noindent\textbf{DIAL \& PAN16}: Next, we focus on the Twitter-based datasets DIAL \& PAN16, where the target task is sentiment classification/mention detection and the protected attribute is one of the demographic associations (race/gender/age) of the author. The evaluation results are reported in Table~\ref{tab:dynamic-results} (third row). For these datasets, we observe that 
(i) ``w/o adversary $h(x)$'' representations have higher F1 and lower MDL for predicting $z$, compared to ``Pre-trained $h(x)$''. This shows that fine-tuning on the target task $y$ encodes information about the protected attribute $z$.
(ii) ``\textsc{AdS} - $h(x)$'' performs similar to ``w/o adversary $h(x)$'' representations on the target task but still leaks significant information about $z$, unlike the previous datasets.
(iii) ``\textsc{AdS} - $s(h(x))$'' achieves the best performance in terms of guarding the protected variable $z$ (achieves almost random performance in PAN16 dataset), without much performance drop in the target task.

\noindent\textbf{Comparison with Prior Work}: We report two metrics following \citet{elazar2018adversarial}: (i) $\Delta_z$ - which denotes the performance above the random baseline for $z$ (50\% for both PAN16 and DIAL) (ii) Acc$_y$ - is the probing accuracy on the target task. 
Our framework cannot be directly compared with \citet{elazar2018adversarial} as they have used LSTM Encoder. Therefore, we report the baseline Encoder performances as well. In Table~\ref{tab:baseline}, 
we observe that it possible to retrieve $z$ and $y$ from ``w/o adversary BERT'' with a higher performance compared to ``w/o adversary LSTM''. This indicates that BERT encodes more information pertaining to both $y$ and $z$ compared to LSTM. 
In the DIAL dataset, \textsc{AdS} is able to reduce $\Delta_z$ by an absolute margin of 25\% compared to 9.7\% by \citet{elazar2018adversarial}, while the absolute drop in Acc$_y$ is 3.5\% compared to 3.6\% by \citet{elazar2018adversarial}. In PAN16 dataset, \textsc{AdS} achieves the best $\Delta_z$ and Acc$_y$ performance for both setups with protected attributes: age and gender respectively.  \textsc{AdS} - $s(h(x))$  also achieves performance comparable  to the ``w/o adversary BERT'' setup, which is fine-tuned on the target task. Therefore, \textsc{AdS} is successful in scrubbing information about $z$ from the representations of a stronger encoder compared to \citet{elazar2018adversarial}.




\begin{table}[t!]
	\centering
	\resizebox{0.5\textwidth}{!}{
        \begin{tabular}{l| c c |c c| c c}
	\hline
	 \multirow{3}{*}{\textsc{Setup}} & \multicolumn{6}{c}{\textsc{Pan16}}\TBstrut\\
	 & \multicolumn{2}{c|}{Age ($z_1$)} & \multicolumn{2}{c|}{Gender ($z_2$)} & \multicolumn{2}{c}{Mention ($y$)}\\
	 & F1$\downarrow$ & MDL$\uparrow$ & F1$\downarrow$ & MDL$\uparrow$ & F1$\uparrow$ & MDL$\downarrow$ \Bstrut\\
	\hline
	Random & 50.0 & - & 50.0 & - & 50.0 & -\Tstrut\\
	w/o adversary $h(x)$ & 66.5 & 196.4 & 69.3 & 192.0 & 88.6 & 6.8\\
	\textsc{AdS} $s(h(x))$ - (age) & 61.5 & 224.2 & 62.6 & 218.7 & {88.7} & 14.3\\
	\textsc{AdS} $s(h(x))$ - (gender) &  60.6 & 222.6 & 64.2 & 216.8 & 88.6 & 12.9\\
	\textsc{AdS} $s(h(x))$ - (both) & \textbf{53.8} & \textbf{231.5} & \textbf{54.4} & \textbf{230.9} & 88.6 & \textbf{5.5}\Bstrut\\
	\hline
\end{tabular}
	}
	\caption{Evaluation results of protecting multiple attributes using \textsc{AdS}. Statistically significant best performances are in bold. Expected trends for a metric are shown in $\uparrow$- higher scores and $\downarrow$- lower scores. ``\textsc{AdS} $s(h(x))$ - (both)'' achieves the best performance.\footnote{\label{ftn:change}Performance of ``\textsc{AdS} - $s(h(x))$'' varies from Table~\ref{tab:dynamic-results} as the dataset size and split in this setup are different.}}
	\label{tab:multi-disc}
\end{table}

\subsection{Scrubbing multiple protected attributes}
\label{sec:multi-disc}
In this experiment, we show that using \textsc{AdS} it is possible to guard information about \textit{multiple} protected attributes. $\mathcal{L}_s$ in this setup is defined as:
\begin{equation*}
    \begin{aligned}
	    \mathcal{L}_s(e_i, y_i) = \mathcal{L}_c(c(u_i), y_i) -\lambda_1 \sum\limits_{n=1}^{N} H(d_n(u_i)) \\+ \lambda_2 \sum\limits_{n=1}^{N} \delta(d_n(u_i))
	\end{aligned}
    \label{eqn:multi-gen-loss}
\end{equation*}

\noindent where $N$ is the number of protected attributes and $d_n(\cdot)$ is the Bias discriminator corresponding to the $n^{th}$ protected attribute $z_n$.

We evaluate on PAN16 dataset considering two protected attributes $z_1$ (age) and $z_2$ (gender). The target task is mention prediction. 
{We consider the subset of PAN16 that contains samples with both gender and age labels. This subset } has 120K training instances and 30K test instances. Evaluation results are reported in Table~\ref{tab:multi-disc}. Similar to previous experiments, we observe that ``w/o adversary $h(x)$'' {(fine-tuned BERT)} leaks information about both protected attributes age and gender. We evaluate the information leak when ``\textsc{AdS} $s(h(x))$'' is retrieved from a setup with single Bias discriminator (age/gender). We observe a significant gain in MDL for the corresponding $z_n$ in both cases, indicating that the respective $z_n$ is being protected. 
Finally, we train \textsc{AdS} using two Bias discriminators and  ``\textsc{AdS} - $s(h(x))$ (both)'' representations achieve the best performance in guarding $z_1$ \& $z_2$, while performing well on the target task. This shows that \textsc{AdS} framework is scalable and can be leveraged to guard multiple protected attributes simultaneously.

\begin{table}[t!]
	\centering
	\resizebox{0.5\textwidth}{!}{
        \begin{tabular}{l| c c c| c  c c}
	\toprule[1pt]
	
	\multirow{2}{*}{Scrubber loss} & \multicolumn{3}{c}{Gender ($z$)} & \multicolumn{3}{c}{Sentiment ($y$)}\TBstrut\\
	& F1$\downarrow$ & P$\downarrow$ & R$\downarrow$ & F1$\uparrow$ & P$\uparrow$ & R$\uparrow$ \Bstrut\\ 
	\midrule[1pt]
	Random & 33.3 & 33.3 & 33.3 & 33.3 & 33.3 & 33.3 \Tstrut\\
	$\delta$-loss (w/o entropy) &  49.5 & 47.7 & 53.9 & 91.2 & 91.2 & 91.2\\
	Entropy (w/o $\delta$-loss) &  35.7 & 36.4 &  53.2 & \textbf{91.5} & \textbf{91.6} & \textbf{91.5}\\
	Entropy + $\delta$-loss &  \textbf{29.8} & \textbf{33.3}  & \textbf{27.0} & 90.2 & 90.5 & 89.9 \Bstrut\\
	\bottomrule[1pt]
\end{tabular}
	}
	\caption{Ablation experiments on Funpedia using F1-score (F1), Precision (P) and Recall (R). Expected trends for a metric are shown in $\uparrow$- higher scores and $\downarrow$- lower scores. \textsc{AdS}  with both loss components performs the best in guarding $z$.}
	\label{tab:ablations}
\end{table}

\subsection{Efficacy of different losses}

We experiment with different configurations of the Scrubber loss $\mathcal{L}_s$ to figure out the efficacy of individual components. We show the experimental results on the Funpedia dataset in Table~\ref{tab:ablations} (with $\lambda_1 = \lambda_2 = 1$). We observe that most leakage in $z$ (increase in prediction F1-score) occur when the entropy loss is removed. Removing $\delta$-loss also results in a slight increase in leakage accompanied by a gain in performance for predicting $y$. This shows that both losses are important for guarding $z$. 

Empirically, we found that $\delta$-loss is not suitable for binary protected attributes. 
{This is because during training when the Scrubber is encouraged to learn representations that do not have information about $z$, it  learns to encode representations in a manner such that the Bias discriminator  predicts the opposite $z$ class.} Hence, the information about $z$ is still present and is retrievable using a probing network $q$. For this reason, we use $\delta$-loss for only Funpedia ($\lambda_2$ values in Table~\ref{tab:setup}) where we considered 3 gender label classes.

\begin{figure}[t!]
	\centering
	
	\subfloat[][ Pre-trained]{
		\includegraphics[width=0.23\textwidth]{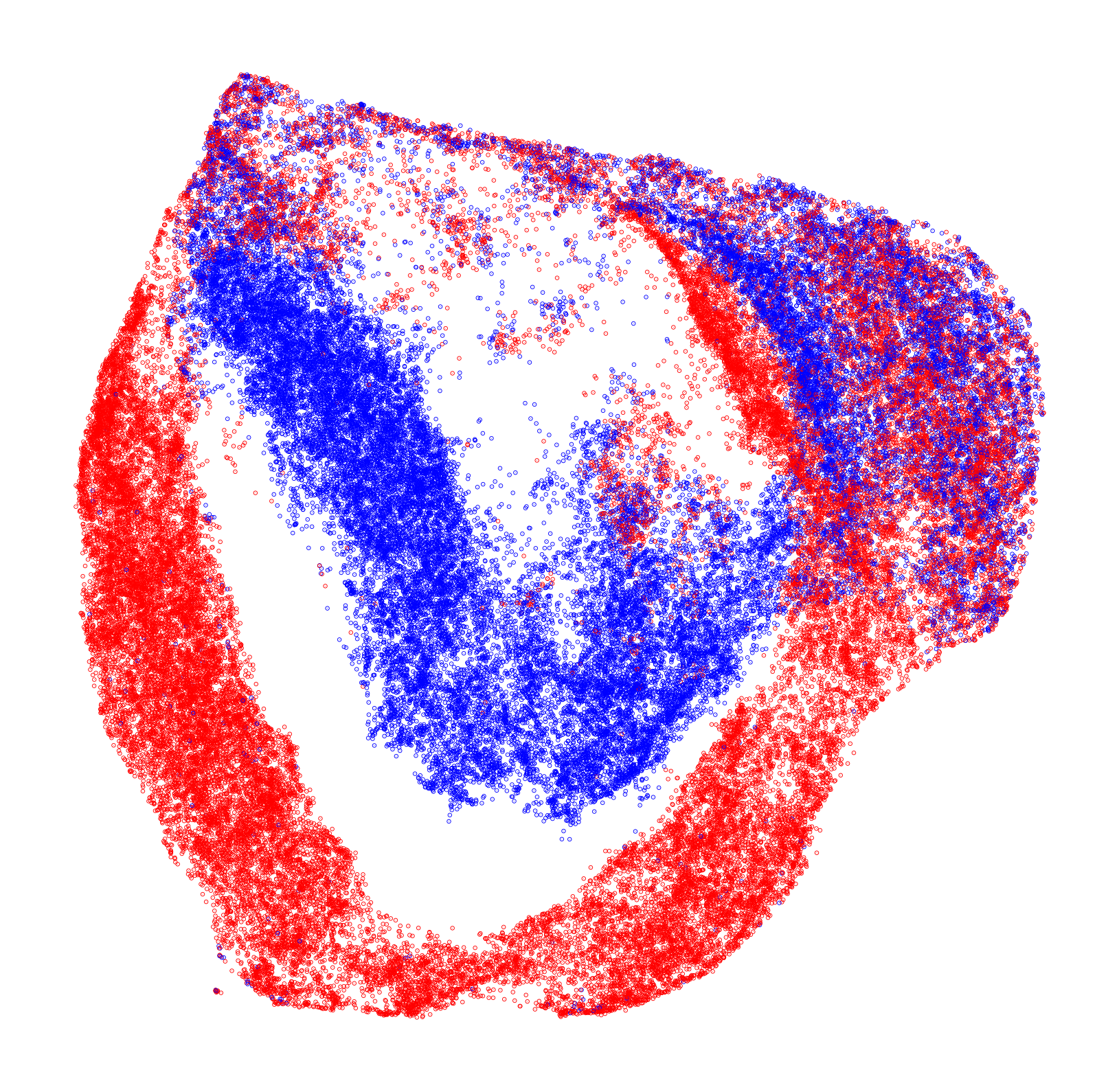}{ 
			\label{subfig:pre-trained}}
	} 
	\subfloat[][ After training]{ %
		\includegraphics[width=0.23\textwidth]{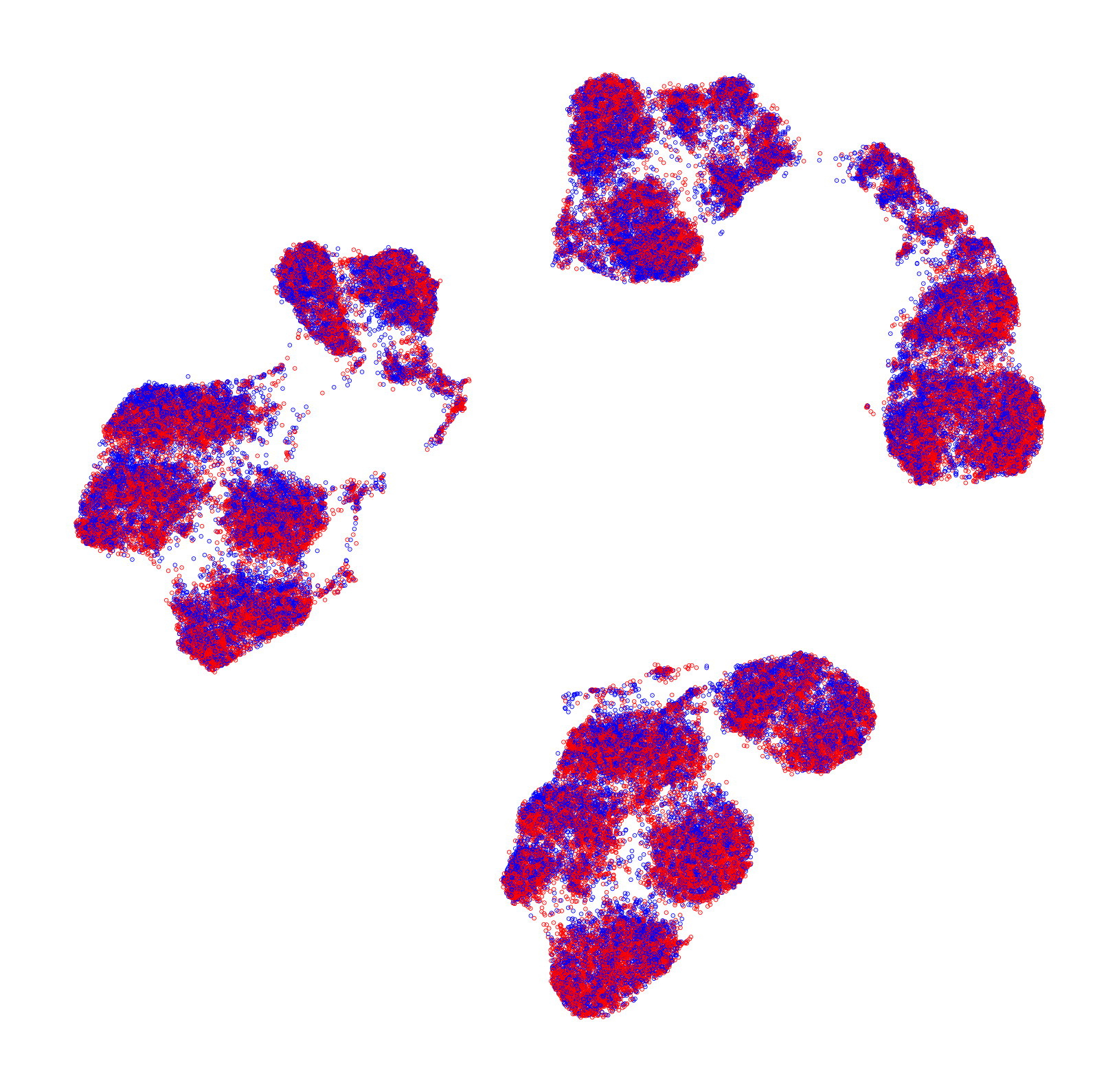}{
			\label{subfig:post-training}} 
	}
	\vspace{5pt}
	\caption{UMAP projection of Scrubber output representations $s(h(x))$ from Biographies corpus with profession as ``professor''. Blue and red labels indicate female and male biographies respectively. (a) Pre-trained BERT representations (b) BERT representations post training in \textsc{AdS}.}
	\label{fig:embedding}
\end{figure}

\subsection{Visualization}

We visualize the UMAP \cite{mcinnes2018umap} projection of Encoder output representations, $h(x)$,  in Figure~\ref{fig:embedding}. {Blue} and {red} labels indicate female and male biographies respectively. {Figure~\ref{subfig:pre-trained} and Figure~\ref{subfig:post-training} show representations before and after \textsc{AdS} training. } In Figure~\ref{subfig:pre-trained}, male and female labeled instances are clearly separated in space. This shows that text representations encode information relating to gender attributes. In Figure~\ref{subfig:post-training}, we observe that after training in our adversarial framework both male and female labeled instances are difficult to segregate. This indicates that post training in \textsc{AdS}, it is difficult to  identify biography representations on the basis of gender.

\section{Conclusion}
In this work, we proposed \textbf{Ad}versarial \textbf{S}crubber (\textsc{AdS}) to remove demographic information from contextual representations. Theoretical analysis showed that under certain conditions, our framework converges without leaking information about protected attributes. We extend previous evaluation metrics to evaluate fairness of representations by using MDL. 
Experimental evaluations on 8 datasets show that \textsc{AdS} is better at protecting demographic attributes than baselines. We show that our approach is scalable and can be used to remove multiple protected attributes simultaneously. 
Future work can explore leveraging \textsc{AdS} towards learning fair representations in other NLP tasks.

\section{Acknowledgement}
This work was supported in part by grants NIH 1R01AA02687901A1 and NSF IIS2133595.

\section*{Ethical considerations}

We propose \textsc{AdS}, an adversarial framework to prevent text classification modules from taking biased decisions. \textsc{AdS} is intended to be used in scenarios, where the user is already aware of the input attributes they want to protect. 
\textsc{AdS} can only be trained on data where protected attributes are annotated. It is possible that representations retrieved from \textsc{AdS}, contain sensitive information which were not defined as the protected variables. Even in such a scenario, \textsc{AdS} won't reveal information more than its already available in the dataset. One potential way of misusing \textsc{AdS} would to define relevant features for a task (e.g. experience for a job application) as a protected attribute, then the classification system may be forced to rely on sensitive demographic information for predictions. In such cases, it is possible to flag systems by evaluating the difference in True Positive Rate (TPR) when the protected attribute is changed ($\mathrm{GAP}^{\mathrm{TPR}}_{z, y}$ metric \cite{de2019bias}). All experiments were performed on publicly available data, where the identity of author was anonymous. We did not perform any additional data annotation.

\bibliography{anthology,custom}
\bibliographystyle{acl_natbib}

\appendix

\section{Appendix}
\subsection{Minimum Description Length}
\label{sec:mdl}


Minimum Description Length (MDL) measures the description length of labels given a set of representations. MDL captures the amount of effort required to achieve a certain probing accuracy, characterizing either complexity of probing model, or amount of data required.

Estimating MDL involves a dataset $\{(x_1, y_1), \ldots, (x_n, y_n)\}$, where $x_i$'s are data representations from a model and $y_i$'s are task labels. Now, a sender Alice wants to transmit labels $\{y_1, \ldots, y_n\}$ to a receiver Bob, when both of them have access to the data representations $x_i$'s. In order to transmit the labels efficiently,  Alice needs to encode $y_i$'s in an optimal manner using a probabilistic model $p(y|x)$. The minimum codelength (\textit{Shannon-Huffman code}), required to transmit  the labels losslessly is: $\mathcal{L}_p(y_{1:n}| x_{1:n}) = - \sum\limits_{i=1}^{n} \log_2 p(y_i | x_i)$.

There are two  ways of evaluating MDL for transmitting the labels $y_{1:n}$ (a) \textit{variational code} - transmit $p(y|x)$ explicitly and  then use it to encode the labels (b) \textit{online code} - encodes the model  and labels without explicitly transmitting the model. In our experiments, we evaluate the online code for estimating MDL. In the online setting, the labels are transmitted in blocks in $n$ timesteps $\{t_0, \ldots, t_n\}$. 
Alice encodes the first block of labels $y_{1:t_1}$ using a uniform code. Bob learns a model $p_{\theta_1}(y|x)$ using the data $\{(x_i, y_i)\}^{t_1}_{i=1}$, Alice then transmits the next block of labels $y_{t_1 + 1:t_2}$ using $p_{\theta_1}(y|x)$. In the next iteration, the receiver trains a new model  using a larger chunk of data $\{(x_i, y_i)\}^{t_2}_{i=1}$, which encodes $y_{t_2 + 1:t_3}$. This continues  till the whole set of labels $y_{1:n}$ is transmitted. The total codelength required for transmission using this setting is given as:

\begin{equation}
\begin{aligned}
	\mathcal{L}_{online}(y_{1:n}|x_{1:n}) = t_1 \log_2 C - \\
	&\hspace{-4.5cm}\sum^{n-1}_{i=1}	\log_2 p_{\theta_i}	(y_{t_i+1:t_{i+1}} |x_{t_i+1:t_{i+1}} )
\end{aligned}
\label{eqn:mdl}
\end{equation}

\noindent where $y_i \in \{1, 2, \ldots, C\}$.  The online codelength $\mathcal{L}_{online}(y_{1:n}|x_{1:n})$ is shorter if the probing model is able to perform well using fewer training instances, therefore capturing the effort needed to achieve a prediction performance.

\subsection{Theoretical Analysis}
\label{sec:prop1}
\textbf{Proposition}. \textit{Minimizing $\delta$-loss is equivalent to increasing the Bias discriminator loss $\mathcal{L}_d$.}

\noindent \underline{Proof}: The $\delta$-loss function can be written as:
\begin{equation}
\begin{aligned}
\delta(o_i) &= m_i^T \mathrm{softmax}_{gumble}(o_i) 
    \\ &= \frac{\exp{(\frac{\log o_i^k + g_k}{\tau})}}{\sum\limits_{j} \exp{(\frac{\log o_i^j + g_j}{\tau})}}
\end{aligned}
\label{eqn:delta-2}
\end{equation}

\noindent where $o_i^j$ is the raw logit assigned to the $j^{th}$ output class, the true output class is $k = z_i$ and $g_j, g_k$  are i.i.d samples from Gumble(0,1) distribution. The cross entropy loss of the bias discriminator $\mathcal{L}_d$ can be written as:

\begin{equation}
    \begin{aligned}
    \mathcal{L}_d = -\log \frac{\exp(o_i^k)}{\sum\limits_j \exp(o_i^j)}
    \end{aligned}
    \label{eqn:Ld}
\end{equation}

The gumble softmax generates a peaked version of the normal softmax distribution. But the individual gumble softmax logit values (Equation~\ref{eqn:delta-2}) are still proportional to vanilla softmax logits (Equation~\ref{eqn:Ld}): $\delta(o_i) \propto \frac{\exp{o_i^k}}{\sum_j \exp{o_i^j}}$. Then, bias discriminator loss  $\mathcal{L}_d$ can be written as:

\begin{equation}
\mathcal{L}_d \propto -\log \delta(o_i)
\label{eqn:propto}
\end{equation}
Therefore, minimizing $\delta(o_i)$ increases $\mathcal{L}_d$.

\begin{table}[t!]
	\centering
	\resizebox{0.33\textwidth}{!}{
        \begin{tabular}{l| c c | c c| c c| c c}
\toprule[1pt]
\multirow{2}{*}{\textsc{Dataset}} & Time/ epoch \Tstrut\\
&  (min.) \\
\midrule[1pt]
\textsc{Funpedia} & 2 \Tstrut\\
\textsc{Wizard} & 1 \\
\textsc{ConvAI2} & 14 \\
\textsc{Light} & 4 \\
\textsc{OpenSub} & 15 \\
\textsc{Biographies} & 260 \\
\textsc{Dial} & 16 \\
\textsc{Pan16} (gender) & 15  \\
\textsc{Pan16} (age) & 15 \\
\bottomrule[1pt]
\end{tabular}
	}
	\caption{Runtime for each dataset.}
	\label{tab:runtime}
\end{table}

\begin{table*}[t!]
	\centering
	\resizebox{0.8\textwidth}{!}{
        \begin{tabular}{l| c c | c c| c c| c c}
\toprule[1pt]

\multirow{2}{*}{\textsc{Dataset}} & \multicolumn{2}{c|}{\shortstack{Pre-trained $h(x)$}} & \multicolumn{2}{c|}{\shortstack{w/o adversary $h(x)$}} & \multicolumn{2}{c|}{\shortstack{\textsc{AdS} $h(x)$}} & \multicolumn{2}{c}{\shortstack{\textsc{AdS} $s(h(x))$}} \TBstrut\\
& $\overrightarrow{\mathrm{MDL}}_{z}$ & $\overrightarrow{\mathrm{MDL}}_{y}$ & $\overrightarrow{\mathrm{MDL}}_{z}$ & $\overrightarrow{\mathrm{MDL}}_{y}$& $\overrightarrow{\mathrm{MDL}}_{z}$ & $\overrightarrow{\mathrm{MDL}}_{y}$ & $\overrightarrow{\mathrm{MDL}}_{z}$ & $\overrightarrow{\mathrm{MDL}}_{y}$   \TBstrut\\
\midrule[1pt]
\textsc{Funpedia} & 1.03 & 1.94 & 1.29 & 0.12 & 1.48 & 0.43 & 1.73 & 0.45\Tstrut\\
\textsc{Wizard} & 1.08 & 2.15 & 1.47 & 0.06 & 1.84 & 0.09 & 1.95 & 0.06\\
\textsc{ConvAI2} & 1.46 & 1.94 & 1.58 & 0.09 & 1.94 & 0.16 & 1.93 & 0.16\\
\textsc{Light} & 1.21 & 2.28 & 1.44 & 0.21 & 1.89 & 0.43 & 1.92 & 0.42\\
\textsc{OpenSub} & 0.92 & 2.03 & 1.49 & 0.12 & 1.77 & 0.18 & 1.78 & 0.18\\
\textsc{Biographies} & 0.11 & 1.94 & 1.74 & 0.01 & 1.73 & 0.01 & 1.74 & 0.01\\
\textsc{Dial} & 1.46 & 1.81 & 1.06 & 0.60 & 1.65 & 0.31 & 1.75 & 0.34\\
\textsc{Pan16} (gender) & 1.87  & 1.62 & 1.67 & 0.03 & 1.90 & 0.04 & 1.96 & 0.05\\
\textsc{Pan16} (age) & 1.89 & 1.64 & 1.85 & 0.03 & 1.89 & 0.03 & 1.97 & 0.04\\
\bottomrule[1pt]
\end{tabular}
	}
	\caption{Probing performance of representations retrieved from different settings in terms of $\overrightarrow{\mathrm{MDL}}$.}
	\label{tab:mdl-norm}
\end{table*}

\subsection{Implementation Details}
\label{sec:impl-details}
All experiments are conducted in PyTorch framework using Nvidia GeForce RTX2080 GPU with 12GB memory. We use an off-the-shelf MLPClassifer from \textit{sklearn}\footnote{https://scikit-learn.org/} as our \textit{probing network} $q$. \textsc{AdS} has a total of 110M parameters (all 4 modules combined). The average runtime per epoch for each dataset is reported in Table~\ref{tab:runtime}.

\subsection{Measuring Fairness in Representations}
MDL scales linearly with the dataset size (Equation~\ref{eqn:mdl}), therefore making it hard to compare across different datasets. In order to make it comparable, we measure a normalized description length measure for transmitting 1000 labels:

\begin{equation}
    \overrightarrow{\mathrm{MDL}} = \frac{1000 \times \mathrm{MDL}}{\lvert\mathcal{D}\rvert}
\end{equation}

\noindent $\lvert\mathcal{D}\rvert$ is the dataset size. 
Performance using this measure are reported in Table~\ref{tab:mdl-norm} for all datasets. In all experiments we report the MDL required for transmitting the labels in the training set.

\label{sec:appendix}

\end{document}